\documentclass[sigconf]{acmart}
\usepackage{multirow}
\usepackage{bigstrut}
\usepackage{bm}
\usepackage{hyperref}
\usepackage{balance}
\AtBeginDocument{%
  \providecommand\BibTeX{{%
    \normalfont B\kern-0.5em{\scshape i\kern-0.25em b}\kern-0.8em\TeX}}}

\copyrightyear{2022}
\acmYear{2022}
\setcopyright{acmcopyright}\acmConference[MM '22]{Proceedings of the 30th ACM International Conference on Multimedia}{October 10--14, 2022}{Lisboa, Portugal}
\acmBooktitle{Proceedings of the 30th ACM International Conference on Multimedia (MM '22), October 10--14, 2022, Lisboa, Portugal}
\acmPrice{15.00}
\acmDOI{10.1145/3503161.3548446}
\acmISBN{978-1-4503-9203-7/22/10}
\acmSubmissionID{2626}

\begin{document}

\title{$T$-former: An Efficient Transformer for Image Inpainting}

\author{Ye Deng}
\orcid{0000-0003-4616-3318}
\affiliation{%
	\institution{Xi'an Jiaotong University}
	\city{Xi'an}
	\state{Shaanxi}
	\country{China}
}
\email{dengye@stu.xjtu.edu.cn}

\author{Siqi Hui}
\affiliation{%
	\institution{Xi'an Jiaotong University}
	\city{Xi'an}
	\state{Shaanxi}
	\country{China}
}
\email{huisiqi@stu.xjtu.edu.cn}

\author{Sanping Zhou}
\authornote{The author is also with the Shunan Academy of Artificial Intelligence, Ningbo, Zhejiang 315000, China.}
\affiliation{%
	\institution{Xi'an Jiaotong University}
	\city{Xi'an}
	\state{Shaanxi}
	\country{China}
}
\email{spzhou@xjtu.edu.cn}

\author{Deyu Meng}
\affiliation{%
	\institution{Xi'an Jiaotong University}
	\city{Xi'an}
	\state{Shaanxi}
	\country{China}
}
\email{dymeng@mail.xjtu.edu.cn}

\author{Jinjun Wang}
\affiliation{%
	\institution{Xi'an Jiaotong University}
	\city{Xi'an}
	\state{Shaanxi}
	\country{China}
}
\email{jinjun@mail.xjtu.edu.cn}

\renewcommand{\shortauthors}{Ye Deng et al.}

\begin{abstract}
Benefiting from powerful convolutional neural networks (CNNs), learning-based image inpainting methods have made significant breakthroughs over the years. However, some nature of CNNs (e.g. local prior, spatially shared parameters) limit the performance in the face of broken images with diverse and complex forms. Recently, a class of attention-based network architectures, called transformer, has shown significant performance on natural language processing fields and high-level vision tasks. Compared with CNNs, attention operators are better at long-range modeling and have dynamic weights, but their computational complexity is quadratic in spatial resolution, and thus less suitable for applications involving higher resolution images, such as image inpainting.  In this paper, we design a novel attention linearly related to the resolution according to Taylor expansion. And based on this attention, a network called $T$-former is designed for image inpainting. Experiments on several benchmark datasets demonstrate that our proposed method achieves state-of-the-art accuracy while maintaining a relatively low number of parameters and computational complexity. The code can be found at \href{https://github.com/dengyecode/T-former_image_inpainting}{github.com/dengyecode/T-former\_image\_inpainting}
\end{abstract}



\begin{teaserfigure}
	\includegraphics[width=\textwidth]{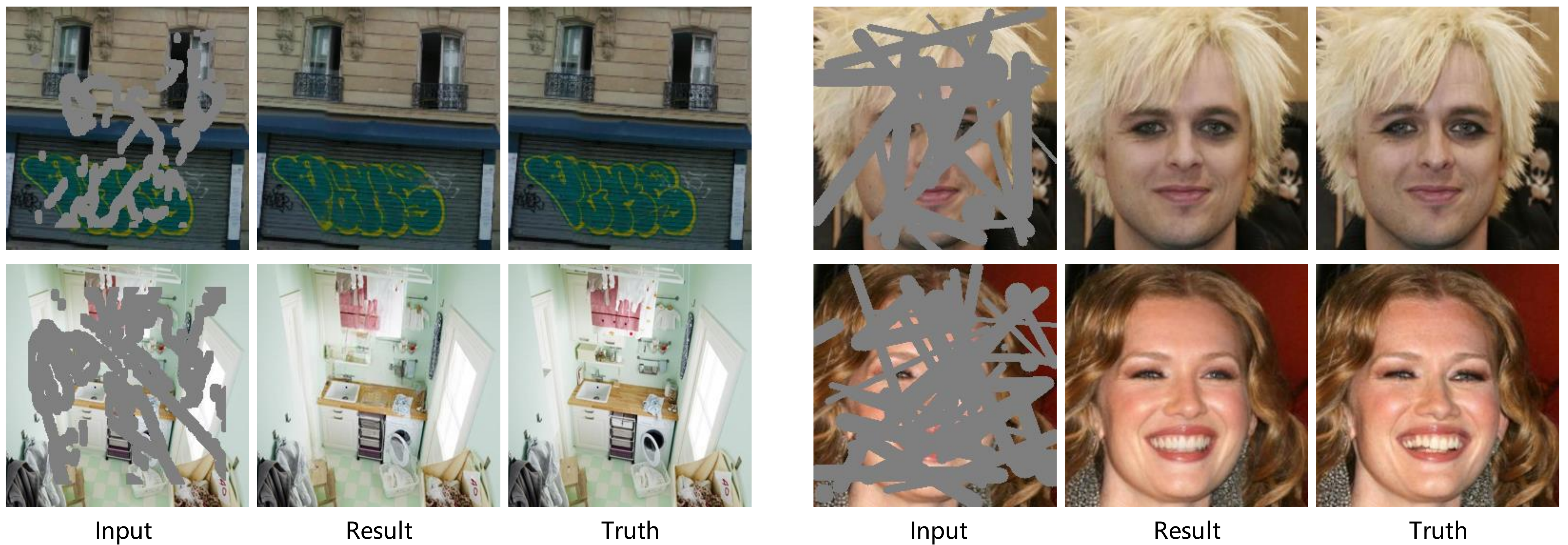}
	\caption{Image inpainting outputs by our proposed $T$-former. In each group, the input image is shown on the left, with gray pixels representing the missing areas. (Best with color and zoomed-in view)}
	\label{fig:teaser}
\end{teaserfigure}

\maketitle

\section{Introduction}
Image inpainting (or completion)~\cite{2000siginpain} is the process of filling in corrupted or missing parts of an image, as Figure \ref{fig:teaser} shown. It is an important task in the field of computer vision and image processing and can benefit users in a wide range of applications, such as removing unwanted objects in image editing.
The key challenge in image inpainting is to make the filled pixels blend in with the non-missing parts.

Prior to deep learning, non-learning inpainting algorithms can be roughly divided into two categories, diffusion-based approaches~\cite{2000siginpain,935036ss,criminisi2004region,bertlamioPDE2006tip} and exemplar-based approaches~\cite{komodakis2007tip,Barnes:2009:PAR,xuzongben2010tip,7056453exempalrinpai}. Diffusion-based approaches smoothly propagate the information from observed boundaries to the interior of damaged areas. However, since diffusion-based approaches do not consider the global image structure, they are only effective in filling small holes and less effective in dealing with large scale breakage. To address the drawbacks of diffusion-based inpainting approaches, the exemplar-based approach searches for valid information from known regions of the entire image and copies or relocates this information to the missing locations. Although the exemplar-based algorithms perform well in the face of simple pattern breakage in larger areas, they do not perform well in filling images with complex patterns because they do not understand the semantic information of the image.

With the development of convolutional neural networks (CNNs), learning-based approaches have reached state-of-arts in the field of image inpainting. These inpainting models~\cite{Pathak_2016_CVPR,IizukaSIGGRAPH2017,PIC_2019CVPR,Lirfr_2020CVPR} formulate the inpainting as a conditional image generation problem and customize a CNNs-based encoder-decoder as their corresponding conditional image generator. By training on sufficiently large scale datasets, CNNs show their strength in learning rich patterns and image semantics, and filling the target regions with such learned knowledge. In addition, the sparse connectivity and parameter sharing of CNNs in space make them computationally efficient. However, some basic characteristics of CNNs make them may have some limitations in handling the inpainting task. (a) the locality. CNNs are good at acquiring local relationships but not good at capturing long-range dependencies. Although the importance of locality for images has been demonstrated in various vision tasks for a long time, for image inpainting, focusing on non-local features (the whole image) is more likely to find the appropriate information for broken regions. (b) spatial-sharing and static parameters. The same convolution kernel operates on features across all spatial locations and the parameters of the kernel are static at the time of inference. This is somewhat inflexible in the face of inpainting tasks where images are mixed with broken and unbroken pixels and the damaged regions are variable.

Recently, (self-)attention~\cite{NIPS2017_3f5ee243}, popular in the field of natural language processing, has been introduced in vision tasks~\cite{dert_eccv2020,dosovitskiy2021an}.
Compared to CNNs, attention operators whose weights dynamically adjust with the input are better able to capture long-range dependencies through explicit interaction with global features. And as a well-explored architecture in language tasks, the transformer model, based on the attention, is emerging in high-level vision tasks. Although the attention operator has advantages over CNNs in some aspects, its computational complexity grows quadratically with spatial resolution and is therefore not particularly suitable for high-resolution images, a situation that occurs frequently in low-level vision tasks including image inpainting. Recently, some designs that can reduce the computational complexity of attention operators have been transferred to inpainting~\cite{dengyemm2021} or other low-level vision tasks~\cite{wang2021uformer,swinlr_2021ICCV}. These methods either apply attention to a sequence of patches unfolded from the image~\cite{dengyemm2021}, or divide the image into non-overlapping parts and compute the attention for each part independently~\cite{wang2021uformer,swinlr_2021ICCV,song2022vision}. However, limiting the spatial extent of attention somewhat defeats the original purpose of capturing the long-range dependence between pixels.

Specifically, for the computational load caused by the dot product and softmax operator in the attention operator, we utilize Taylor's formula to approximate exponential function, and then reduce the computational complexity by swapping the computational order of matrix multiplication. In addition, to mitigate the performance loss caused by the error in the Taylor approximation, we introduced the gating mechanism~\cite{gatedlinear_icml2017} for the proposed attention operator. The previous work~\cite{yuGC_2019ICCV} showed that the gating mechanism on the convolution in the inpainting can be seen as controlling which features should flow forward. The gating mechanism we impose on the attention is equivalent to an adjustment of the "inaccurate" attention, allowing the subsequent layers in the network to focus on the information that will help the inpainting, thus producing a high quality complementation result.

In this paper, based on our designed linear attention, we propose an U-net~\cite{Unet2015} style network, called $T$-former, for image inpainting. Compared with the convolution-based encoder-decoder, in $T$-former we replace the convolution with the designed transformer module based on the proposed linear attention. Our proposed $T$-former combines the texture pattern learning capability of CNNs with the ability of the attention to capture long-range dependencies, and the complexity of this attention is linear rather than quadratically related to the resolution. Our proposed $T$-former is able to achieve performance comparable to other advanced models while maintaining a small complexity compared to those models.
 \section{Related Work}
 \subsection{Vision Transformer}
The transformer model~\cite{NIPS2017_3f5ee243} is a neural network centered on the (self-)attention that plays an important role in natural language processing, and  Carion \emph{et al.}~\cite{dert_eccv2020} were the first to introduce it into the field of vision for object detection. 
Dosovitskiy \emph{et al.}~\cite{dosovitskiy2021an} then designed a transformer structure more suitable for use in the visual field based on the characteristics of images. Touvron \emph{et al.}~\cite{pmlr-deit-touvron21a} reduced the data requirements of visual transfoemer with the help of knowledge distillation. Wang \emph{et al.}~\cite{Wang_2021_ICCV_pvt} then introduced the feature pyramid idea commonly used to build CNNs networks into transformer network construction, which improved the performance of transformer networks. Next, Vaswani~\emph{et al.}~\cite{Vaswani_2021_CVPR} reduce the computational demand of the model by limiting the range of attention so that the self-attention acts only on a local window. Subsequently, Liu \emph{et al.}~\cite{Liu_2021ICCVswin} extended the use and performance of the transformer model by more subtle design of window attention. These works demonstrated the potential of the transformer for high-level vision tasks, yet because its core self-attention excels in features such as long-range modeling, it also meets the needs of low-level tasks such as inpainting.  However, the computational complexity of the attention in the transformer grows quadratically with spatial resolution, making it inappropriate for direct use in low-vision tasks that require the generation of higher-resolution outputs. Therefore, a class of models chooses to process only low-resolution features of the image with transformer. VQGAN~\cite{vqgan2021CVPR}, an autoregressive transformer is utilized to learn the effective codebook. ImageBART~\cite{nips2021_imagebart} improves the quality of image synthesis by replacing the autoregressive model in VQGAN with the diffusion process model. MaskGIT~\cite{Cmaskgit_2022CVPR}, in contrast to VQGAN, abandons the autoregressive generation paradigm and introduces a mask, which determines the inference token by the probability value of the mask instead of synthesizing it sequentially as in autoregressive. ICT\cite{Wan_2021_ICCVdiverseT} is a two-stage inpainting where the first stage gets a coarse result by transformer and then feeds this result into a CNN to refine the details. 
BAT~\cite{yumm2021diverse} improves on the first stage of ICT by introducing bidirectional and autoregressive transformer to improve the capability of the model. 
TFill~\cite{tfill_2022CVPR} introduces a restricted CNN head on the transformer in ICT to mitigates the proximity influence. These approaches allowed the models to obtain more compact image encodings, but still did not change the limitation that they could not be applied to high resolution images. Subsequently, a different strategy to reduce the complexity was generally adopted.
For example, Zamir \emph{et al.}~\cite{zamir2021restormer} propose replacing spatial attention with inter-channel attention. Or the replacement of inter-pixel attention with inter-patch attention as in  \cite{zeng_2020eccv_stt,dengyemm2021}. There is also the use of the window attention as in~\cite{wang2021uformer,swinlr_2021ICCV} to reduce computational complexity by directly limiting the spatial range of action of attention in a similar way to \cite{Liu_2021ICCVswin}. Our $T$-former, which does not avoid the problem of attention between full-space pixels, learns long-range dependencies without imposing excessive complexity.

 \subsection{Image Inpainting}
 Prior to deep learning, non-learning methods could only fill pixels based on the content of the missing regions around~\cite{2000siginpain,935036ss,criminisi2004region,bertlamioPDE2006tip} or all observed regions~\cite{komodakis2007tip,Barnes:2009:PAR,xuzongben2010tip,7056453exempalrinpai} because they could not understand the semantics of the image. These methods tend to be more effective for small missing holes or simple background filling, and have limited effect in the face of images with complex patterns.
 In order to enable the model to output semantic results, Pathak \emph{et al.}~\cite{Pathak_2016_CVPR} introduced the generative adversarial network (GAN)~\cite{NIPS2014_5ca3e9b1} framework to  train a conditional image generation model with the help of convolutional neural networks (CNNs). Then, in response to the shared, static parameters of the convolution, some researchers have modified the convolution so that it can manually~\cite{pc_2018ECCV} or automatically~\cite{yuGC_2019ICCV,XieBA_2019ICCV} adjust the features according to the image breakage.
Next, since it is not easy for the model to recover complex patterns directly, some researchers have chosen to guide the model to complete the image with the help of additional extra image information (e.g., edges~\cite{Nazeri_2019_ICCVW}, structure~\cite{Listrcutre_2019ICCV,Ren_2019_ICCV,liufe_2020eccv,Guodts_2021ICCV}, semantics~\cite{liao2020guidance,Liao_2021CVPR_semantic}).
To improve this, the researchers designed a class of attention operators called contextual attention~\cite{Yan_2018_ECCV,Yuca_2018CVPR,wangijcai2019msa,Liu2019ICCV_csa,pyramid_2019CVPRzeng}. Specifically, with the help of the attention module, they explicitly search the entire image for appropriate content to fill the missing regions. Nonetheless, the high burden of performing attention limits its large-scale deployment in the network, so the model is limited in the extent to which it can improve its long-range modeling capabilities as well as its complementary quality. 
In contrast, our proposed linear attention in $T$-former is not only able to model long-range dependencies between features, but also reduces the complexity compared to the vanilla attention. This enables us to deploy more attention operators in the proposed $T$-former and achieve state-of-the-art in image inpainting.

\section{Approach}
The goal of image inpainting is to fill the target area of the input image $I_{m} \in \mathbb{R}^{C \times H \times W}$ with the appropriate pixels so that the image looks intact. To achieve this goal, we designed an U-net~\cite{Unet2015} style network, based on our proposed linear attention module. In this section, we present our approach from bottom to top. We first describe our proposed linear attention module, and then introduce the architecture of our inpainting network.
\subsection{Linear Attention}
\paragraph{Vanilla Attention} We first explain why the attention operator of the vanilla transformer~\cite{NIPS2017_3f5ee243} model is not applicable to images with higher resolution. Considering a feature map $X \in \mathbb{R}^{C \times H \times W}$, assuming $N = H \cdot W$, the attention operator first feeds the feature $X$ through three different transformations and reshapes them into the two-dimensional matrix to obtain the corresponding three embeddings: the query $Q = [q_1, q_2, \cdots, q_N]^\top \in \mathbb{R}^{N \times C}$, the key $K =[k_1, k_2, \cdots, k_N]^\top \in \mathbb{R}^{N \times C}$, and the value $V = [v_1, v_2, \cdots, v_N]^\top \in \mathbb{R}^{N \times C}$. Then the corresponding attention result $O\in \mathbb{R}^{N \times C}$ can be obtained by:
\begin{equation}
		\begin{split}
		O &= [o_1, o_2, \cdots, o_3]^\top  \\
	  &=\mathcal{A}\left(X \right) \\
	  &= \operatorname{Softmax}\left(Q K^\top / \sqrt{C} \right)V
	  	\end{split}
\end{equation}
where $\mathcal{A}(\cdot)$ the attention function which has quadratic space and time complexity with respect to $N$. And each $o_i \in \mathbb{R}^{C}$ can be obtained as:
\begin{equation}
		o_i= \sum_{j=1}^{N} \frac{\exp{\left(q_i k_j^\top / \sqrt{C} \right)}}{\sum_{l=1}^N \exp{\left(q_i k_l^\top / \sqrt{C}\right)}}v_j \label{eq:dpa}
\end{equation}
The above equation is the dot-product attention with softmax normalization. We can find that the complexity of computing each row $o_i$ in $O$ is $\mathcal{O}(NC)$. Therefore, the computational complexity of $O$ is obtained as $\mathcal{O}(N^2C) = \mathcal{O}((HW)^2C)$, which is quadratic with respect to the image resolution $HW$.

\paragraph{Linearization of Attention}
We can notice that the computational complexity of Eq.~(\ref{eq:dpa}) mainly comes from the softmax term, therefore most linearizations of attention focus mainly on modifications to softmax. Revisiting Eq.~(\ref{eq:dpa}), previous methods~\cite{peng2021random,zhen2022cosformer,pmlr-v119-katharopoulos20a} compute the attention by using different kernel functions $K(q, k)$ instead of $\exp(qk^T)$, by: \begin{equation}
	\begin{split} \label{eq:linearother}
	o_i &= \sum_j \frac{K(q_i, k_j)}{\sum_l K(q_i, k_l)}v_j  = \sum_j \frac{f(q_i) f(k_j)^\top v_j}{\sum_l f(q_i) f(k_l)^\top} \\
	 &= \frac{f(q_i) \sum_j  (f(k_j)^\top v_j )}{\sum_j f(q_i) f(k_j)^\top} 
	 \end{split}
\end{equation}
Note that the property of kernel function $K(q, k)=f(q) f(k)^\top$ is used here, and $f(\cdot)$ is a projection. These methods obtain linear attention ($\mathcal{O}(HWC^2)$) by changing the order of computation of matrix multiplication from $qk^\top v$ to $q(k^\top v)$.

Inspired by the above linear attention approaches, in this paper we take another perspective to linearize the attention by approximating the exponential function through Taylor expansion.
Specifically, we note that Taylor's formula of the exponential function constituting the softmax operator is:\begin{equation}
	\exp(x) = e^x \approx 1 + x  \label{eq:tl}
\end{equation}
Putting Eq.~(\ref{eq:tl}) into Eq.~(\ref{eq:dpa}), we can get (the channel $C$ is ignored for simplicity):\begin{equation}
	\begin{split}
	o_i &=\sum_{j=1}^{N} \frac{\exp{(q_i k_j^\top)}}{\sum_{l=1}^N \exp{(q_i k_l^\top)}}v_j \\
	&= \sum_{j=1}^{N} \frac{1+ q_i k_j^\top}{\sum_{l=1}^N (1 + q_i k_l^\top)}v_j \\
	&= \sum_{j=1}^{N} \frac{v_j+ q_i k_j^\top v_j}{n + q_i\sum_{l=1}^N k_l^\top}  \\
	&= \sum_{j=1}^{N} \frac{v_j+ q_i (k_j^\top v_j)}{n + q_i\sum_{l=1}^N k_l^\top} 
	\end{split} \label{eq:linears} 
\end{equation}
It is worth noting that the last line in Eq.~\ref{eq:linears} is obtained by the properties of vector multiplication.
\begin{figure*}[ht]
	\centering
	\includegraphics[width=1\textwidth]{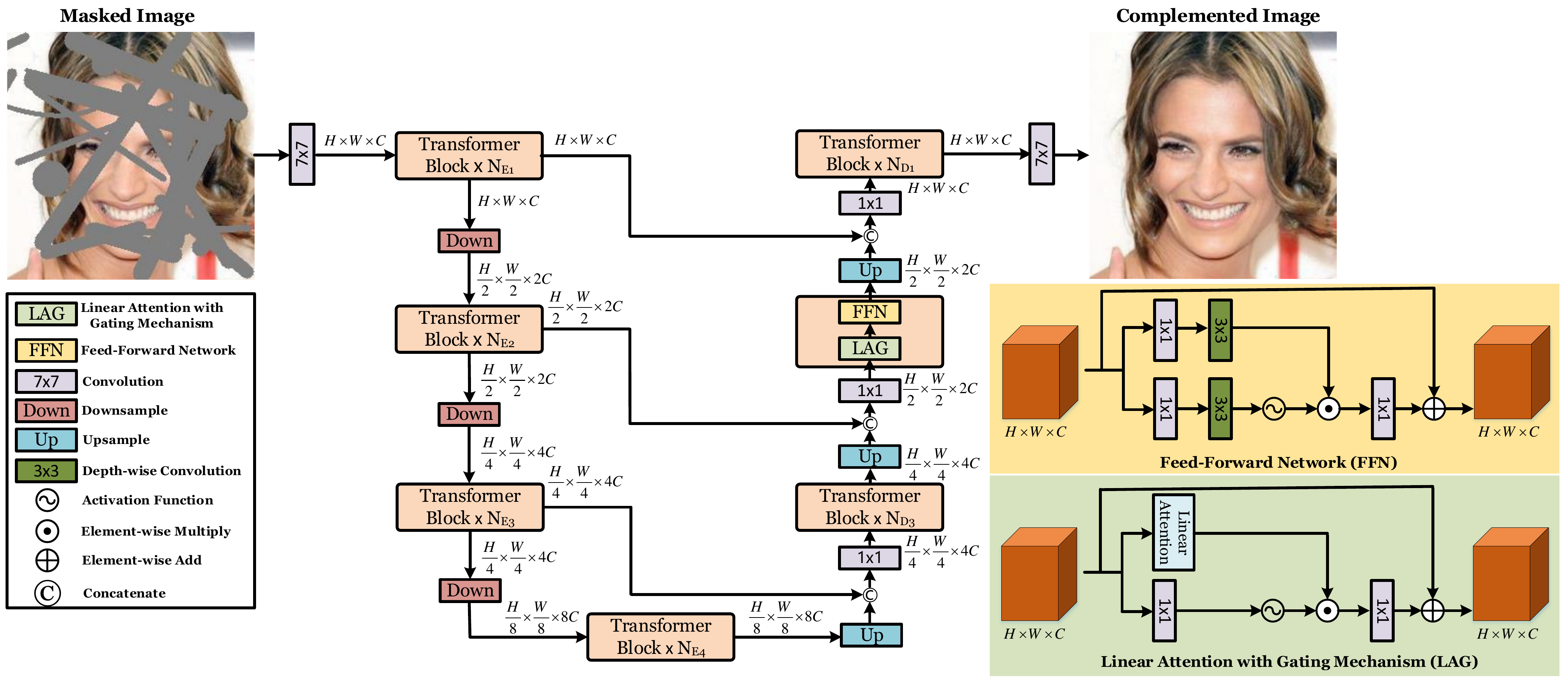} 
	\caption{Overview of our proposed $T$-former. Our model accepts masked images as input and outputs complemented images. Our $T$-former which is an U-net style network composed of transformer blocks that we designed. The transformer block we designed contains two sublayers: (1) Linear attention with gating mechanism (LAG) that performs our proposed linear attention for full-space feature interaction, supplemented with a gating mechanism; (2) Feed-forward network (FFN) that transforms the features learned by the attention operator to send useful representations for subsequent layers.}
	\label{fig:model}
\end{figure*}
\paragraph{Analysis}
From the above, a linear complexity version of attention can be obtained 
from the properties of matrix multiplication: 
\begin{equation}
	V + \left( Q K^\top \right) V  = V + Q  \left(K^\top V \right) \label{eq:matrx}
\end{equation}
In Eq.~(\ref{eq:matrx}), instead of calculating the attention matrix $A = QK^\top \in  \mathbb{R}^{N \times N}$ first, $K^\top V \in \mathbb{R}^{C \times C}$ is computed first and then multiplying $Q \in \mathbb{R}^{N \times C}$. With the help of this trick, the computational complexity of the attention operation is $\mathcal{O}(NC^2) = \mathcal{O}(HWC^2)$. It is noted that in the task of image inpainting, the feature (channel) dimension $C$ is always much smaller than the spatial resolution $H \times W$, so we reduce the computational complexity of the model by a large amount. Also similar to the vanilla transformer~\cite{NIPS2017_3f5ee243}, we also use a multi-headed~\cite{NIPS2017_3f5ee243} version of attention to enhance the performance of our proposed linear attention operator.
Furthermore, the term $V +$ seems to be seen as a residual term with respect to $Q K^\top V$, and from the ablation experiments (as seen in Table~\ref{tab:exp:abl}) we find that it improves the performance of our inpainting model.

\subsection{Gated Mechanism for Linear Attention}
The gating mechanism from recurrent neural networks (GRU~\cite{cho2014properties}, LSTM~\cite{hochreiter1997long}) initially proved its effectiveness on language models~\cite{gatedlinear_icml2017}. And the gating mechanism is also widely used in the feed-forward networks (FFN) of the state-of-arts transformer networks~\cite{shazeer2020glu,hua2022transformer,zamir2021restormer}. A gating mechanism, (or gated linear unit) can be thought of as a neural network layer whose output $O$ is the product of the components of two linear transformations of the input $X$, as: \begin{equation}
	O = I \odot G \quad	I = \phi_i (W_uX)   \quad  G = \phi_g (W_g X)  \label{eq.gate} 
\end{equation}
where $W_i$, $W_j$ are the learnable parameters, $\phi_i$, $\phi_j$ are the corresponding activation functions (which can be absent), and $\odot$ denotes the Hadamard product. The simple and effective gating mechanism significantly enhances the performance of the network making us want to generalize it to the proposed linear attention operator.
Specifically, for an input feature $X$ and the linear attention operator $\mathcal{A}(\cdot)$, then the out $O$ of the attention with a gating mechanism can be written as: \begin{equation}
	O = A \odot G \quad	A = \mathcal{A}(X) \quad G = \phi_g (W_g X)  \label{eq.attn} 
\end{equation}
The gating mechanism applied on the convolution~\cite{yuGC_2019ICCV} plays an important role in the field of image inpainting and can be seen as distinguishing invalid features caused by broken pixels in the input image. Since our proposed linear attention is an ``inaccurate'' attention, we complement our linear attention with a gating mechanism that allows subsequent layers in the network to focus on features that contribute to the inpainting.

\subsection{Network Architecture}
Our $T$-former is an U-net~\cite{Unet2015} style network based on the proposed transformer block, containing both encoder-decoder parts, as shown in Figure \ref{fig:model}. The design of this transformer block we refer to the encoder block in vanilla transformer~\cite{NIPS2017_3f5ee243} and contains two sub-layers.  The first is the proposed linear attention with gating mechanism (LAG), and the second is a simple feed-forward network (FFN). In addition, we adopt a residual connection~\cite{resnet_2016CVPR} adhering to each of the sub-layers. Besides these transformer modules, we also use some convolutional layers to cope with scale changes (like upsampling) of features (inputs). 
\paragraph{Encoder Pipeline.} 
Given a masked images $I_m \in \mathbb{R}^{3 \times H \times W}$, our encoder part first feeds it into a $7 \times 7$ convolution layer and then get the corresponding feature map $E_0 \in \mathbb{R}^{C \times H \times W}$. Here $H$ and $W$ represent the dimension of the spatial resolution and $C$ denotes the dimension of the channel. Next these features are fed into 4-level encoder stages. Each level stage contains a stack of the designed transformer blocks, and we use a convolution with kernel size $3 \times 3$ and stride 2 to downsample the features between every two stages. For the given feature map $E_0 \in \mathbb{R}^{C \times H \times W}$, the $i$-level encoder stage of the transformer block produces the feature map $E_i \in \mathbb{R}^{2^{i-1}C \times \frac{H}{2^{i-1}} \times \frac{W}{2^{i-1}}}$. By the way the feature map output by the final (4-level) stage of the encoder is $E_4 \in \mathbb{R}^{8C \times \frac{H}{8} \times \frac{W}{8}}$.
\paragraph{Decoder Pipeline.} 
The decoder takes the final feature map of encoder $E_4$ as input and progressively restores the high resolution representations. The decoder part consists of 3-level (arranged from largest to smallest) stages, each of which is stacked by several transformer blocks. In each stage of the decoder, the features are first passed through an upsampling layer consisting of nearest neighbor interpolation and $3 \times 3$ convolution.  Given a feature map, the $i$-level decoder stage of the upsampling produces the feature map $D_i \in \mathbb{R}^{2^{i-1}C \times \frac{H}{2^{i-1}} \times \frac{W}{2^{i-1}}}$. 
In addition, to help the decoder part, the encoder feature $E_{i}$ are concatenated to the decoder feature $D_i$ via a skip connection. And the concatenation operation is followed by a $1 \times 1$ convolution layer to decrease the channels (by half). These fused features are then fed into the corresponding transformer block to obtain the dimensionally invariant features $\bar{D}_i$.
Finally, after last (1-level) stage of the decoder we add a $7 \times 7$ convolution layer to convert the feature map $\bar{D}_1 \in \mathbb{R}^{C \times H \times W} $ into the complemented image $I_{out} \in \mathbb{R}^{3 \times H \times W}$. 
\paragraph{Transformer Block.}
As shown in the Figure \ref{fig:model}, the transformer block we used in $T$-former contains two sub-layers.  The first is the proposed linear attention with the gating mechanism (LAG), and the second is a simple feed-forward network (FFN). In the LAG layer, the gating value we obtain by feeding the input $X$ into a $1 \times 1$ convolution layer with a GELU~\cite{hendrycks2016gelu} activation function, i.e. $\phi_g (\cdot)$ and $W_g$ of Eq.~(\ref{eq.attn}).

For the design of the FFN we refer to the recent transformers~\cite{zamir2021restormer,hua2022transformer}, which uses a gate-linear layer~\cite{gatedlinear_icml2017} with residual connections~\cite{resnet_2016CVPR} instead of a residual block~\cite{resnet_2016CVPR} composed of two convolutions in series. Specifically, to reduce the complexity, for the input $X \in \mathbb{R}^{C \times H \times W}$, whose parameter $W_i$, $W_j$ in Eq.~(\ref{eq.gate}) we replace the standard convolution with a combination of a $1 \times 1$ convolution and a $3 \times 3$ depth-wise convolution. 

\subsection{Loss Function}
The loss function $L$ used to train our $T$-former can be written as:
\begin{equation}
	L=\lambda_{r} L_{\mathrm{re}}  + \lambda_{p} L_{\mathrm{perc}} +\lambda_{s} L_{\mathrm{style}} + \lambda_{a} L_{\mathrm{adv}} 
\end{equation}
where $L_{\mathrm{re}}$ represents the reconstruction Loss, $L_{\mathrm{perc}}$ denotes the perceptual loss \cite{johnson2016perceptual}, $L_{\mathrm{style}}$ denotes the style loss \cite{styloss_2016CVPR} and $L_{\mathrm{adv}}$ is the adversarial loss \cite{NIPS2014_5ca3e9b1}. And we set $\lambda_{r}=1$, $\lambda_{p}=1$, $\lambda_{s}=250$, and $\lambda_{a}=0.1$. We will describe each loss function in detail below

\paragraph{Reconstruction Loss}
The reconstruction loss $L_{\mathrm{re}}$ refers to the $L_1$-distance between the output $I_{out}$ and the ground truth $ I_{g}$, which can be defined as:
\begin{equation}
	L_{\mathrm{re}} = \|I_{out} - I_{g} \|_1
\end{equation} 

\paragraph{Perceptual Loss}
The perceptual loss $L_{\mathrm{perc}}$ is formulated with:
\begin{equation}
	L_{\mathrm{perc}}=\mathbb{E}\left[\sum_{i} \frac{1}{N_{i}}\left\|\phi_{i}\left(\boldsymbol{I}_{out}\right)-\phi_{i}\left(\boldsymbol{I}_{g}\right)\right\|_{1}\right]
\end{equation}
where $\phi_i$ is the activation function of the $i$-th layer of the VGG-19~\cite{vgg2014iclr} pre-trained on ImageNet~\cite{deng2009imagenet}.  
\paragraph{Style Loss}
If the size of feature maps is $C_j \times H_j \times W_j$, then the style loss $L_{\mathrm{style}}$
is calculated by:\begin{equation}
	L_{\mathrm{style}}=\mathbb{E}_{j}\left[\left\|G_{j}^{\Phi}\left(\boldsymbol{I}_{out}\right)-G_{j}^{\Phi}\left(\boldsymbol{I}_{g}\right)\right\|_{1}\right]
\end{equation}
Where $G_{j}^{\Phi}$ denotes a $C_j \times C_j$ Gram matrix constructed by the corresponding activation maps $\phi_{j}$. 
\paragraph{Adversarial Loss}
The adversarial loss $L_{\mathrm{adv}}$ is formulated with:\begin{equation}
	L_{\mathrm{adv}}=\mathbb{E}_{\boldsymbol{I}_{g}} \left[\log D\left(\boldsymbol{I}_{g}\right)\right] 
	+\mathbb{E}_{\boldsymbol{I}_{out}}  \log \left[1-D\left(\boldsymbol{I}_{out}\right)\right]
\end{equation}
where $D$ represents a patch GAN discriminator \cite{patchgan_iccv2017} with the spectral normalization~\cite{miyato2018spectral}.

\begin{table*}[ht]
	\centering
	\caption{Numerical comparisons on the several datasets. The $\downarrow$ indicates lower is better, while $\uparrow$ indicates higher is better}
				\resizebox{\textwidth}{!}{
	\begin{tabular}{c|c|cccc|cccc|cccc}
		\toprule
		\multicolumn{2}{c|}{DataSet} & \multicolumn{4}{c|}{Paris Street View} & \multicolumn{4}{c|}{Celeba-HQ} & \multicolumn{4}{c}{Places2} \\
		\midrule
		\multicolumn{2}{c|}{Mask Ratio} & 10-20\% & 20-30\% & 30-40\% & 40-50\% & 10-20\% & 20-30\% & 30-40\% & 40-50\% & 10-20\% & 20-30\% & 30-40\% & 40-50\% \\
		\midrule
		\multirow{5}[2]{*}{FID$\downarrow$} & GC    & 20.68 & 39.48 & 58.66 & 82.51 & 2.54  & 4.49  & 6.54  & 9.83  & 18.91 & 30.97 & 45.26 & 61.16 \\
		& RFR   & 20.33 & 28.93 & 39.84 & 49.96 & 3.17  & 4.01  & 4.89  & 6.11  & 17.88 & 22.94 & 30.68 & 38.69 \\
		& CTN   & 18.08 & 24.04 & 36.31 & 48.46 & 1.77  & 3.33  & 5.24  & 7.69  & 15.70 & 26.41 & 40.05 & 55.41 \\
		& DTS   & 16.66 & 31.94 & 47.30 & 65.44 & 2.08  & 3.86  & 6.06  & 8.58  & 15.72 & 27.88 & 42.44 & 57.78 \\
		& Ours  & \textbf{12.15} & \textbf{22.63} & \textbf{34.47} & \textbf{46.60} & \textbf{1.40} & \textbf{2.55} & \textbf{3.88} & \textbf{5.42} & \textbf{10.85} & \textbf{17.96} & \textbf{26.56} & \textbf{34.52} \\
		\midrule
		\multirow{5}[2]{*}{PSNR$\uparrow$} & GC    & 32.28 & 29.12 & 26.93 & 24.80 & 32.25 & 29.10 & 26.71 & 24.78 & 28.55 & 25.22 & 22.97 & 21.24 \\
		& RFR   & 30.18 & 27.76 & 25.99 & 24.25 & 30.93 & 28.94 & 27.11 & 25.47 & 27.26 & 24.83 & 22.75 & 21.11 \\
		& CTN   & 31.22 & 28.62 & 26.62 & 24.91 & 32.84 & 29.75 & 27.35 & 25.41 & 27.83 & 24.91 & 22.83 & 21.18 \\
		& DTS   & 32.69 & 29.28 & 26.89 & 24.97 & 32.91 & 29.51 & 27.02 & 25.13 & 28.91 & 25.36 & 22.94 & 21.21 \\
		& Ours  & \textbf{32.79} & \textbf{29.72} & \textbf{27.47} & \textbf{25.47} & \textbf{33.36} & \textbf{30.15} & \textbf{27.67} & \textbf{25.67} & \textbf{29.06} & \textbf{25.69} & \textbf{23.36} & \textbf{21.52} \\
		\midrule
		\multirow{5}[2]{*}{SSIM$\uparrow$} & GC    & 0.960 & 0.925 & 0.872 & 0.800 & 0.979 & 0.959 & 0.931 & 0.896 & 0.944 & 0.891 & 0.824 & 0.742 \\
		& RFR   & 0.943 & 0.908 & 0.861 & 0.799 & 0.970 & 0.958 & 0.939 & 0.913 & 0.929 & 0.891 & 0.830 & 0.756 \\
		& CTN   & 0.955 & 0.921 & 0.872 & 0.812 & 0.981 & 0.964 & 0.940 & 0.909 & 0.942 & 0.892 & 0.827 & 0.746 \\
		& DTS   & 0.963 & 0.929 & 0.875 & 0.812 & 0.981 & 0.962 & 0.937 & 0.905 & 0.952 & 0.901 & 0.834 & 0.755 \\
		& Ours  & \textbf{0.964} & \textbf{0.933} & \textbf{0.887} & \textbf{0.825} & \textbf{0.983} & \textbf{0.967} & \textbf{0.945} & \textbf{0.915} & \textbf{0.953} & \textbf{0.907} & \textbf{0.846} & \textbf{0.770} \\
		\bottomrule
	\end{tabular}}%
	\label{tab:exp_all}%
\end{table*}%

\begin{table}[htbp]
	\centering
	\caption{Complexity measure of different models. Including multiply–accumulate operation count (MAC) and number of parameters (Params). Compared to other baseline models, our $T$-former has a smaller number of parameters and computational complexity}
	\resizebox{0.46\textwidth}{!}{
		\begin{tabular}{c|ccccc}
			\toprule
			Model & GC    & RFR   & CTN   & DTS  &  Ours \\
			\midrule
			MAC   & 103.1G & 206.1G & 133.4G & 75.9G & 51.3G \\
			Params & 16.0M & 30.6M & 21.3M & 52.1M & 14.8M \\
			\bottomrule
	\end{tabular}}%
	\label{tab:com}%
\end{table}%

\section{Experiments}
We evaluated our proposed $T$-former on three datasets, including Paris street view (Paris) \cite{Pathak_2016_CVPR}, CelebA-HQ \cite{pgancelebahq_iclr2018} and Places2~\cite{places2tpami2018}. For CelebA-HQ, we use the first 2000 images for test
and the rest for training. For Paris and Places2, we follow the training, testing, and validation splits themselves. During the experiments, all images in datasets were resized to $256 \times 256$.  Furthermore, during the experiments in image inpainting we have to specify the location of the broken areas. Therefore, we use the mask dataset from the PC~\cite{pc_2018ECCV} to simulate the location of the corruption.
The $T$-former we propose was based on a Pytorch~\cite{NEURIPS2019pytorch} implementation and was trained on one RTX3090 (24 GB) with a batch size of 6.  From input to output, the number of transformer blocks of different levels is 1, 2, 3, 4, 3, 2, 1 in order. 
We used the AdamW~\cite{loshchilov2018adamw} optimizer to train the model with a learning rate of $10^{-4}$ and then fine-tune the model with a learning rate of $10^{-5}$. Specifically, on the CelebA-HQ and Paris street view we trained 450,000 iterations and then fine-tuned 200,000 iterations. As for the Places2 data set, we trained about 1 million iterations and then fine-tuned 500,000 iterations. 
\begin{figure*}[htbp]
	\centering
	\includegraphics[width=0.95\textwidth]{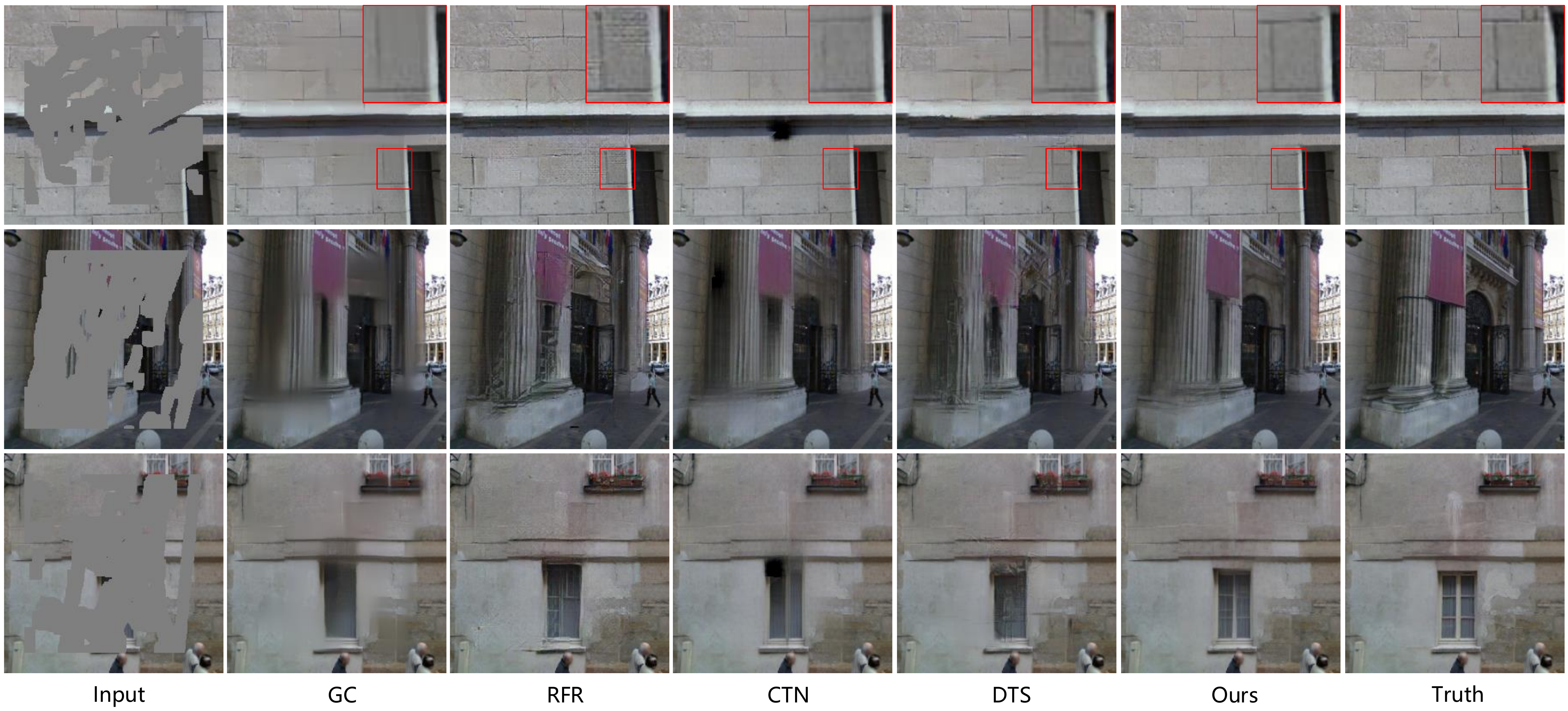} 
	\caption{Qualitative results on the Paris with GC \protect\cite{yuGC_2019ICCV}, RFR \protect\cite{Lirfr_2020CVPR}, CTN \protect\cite{dengyemm2021}, DTS \protect\cite{Guodts_2021ICCV} and our $T$-former. (Best viewed with zoom-in)}
	\label{fig:paris}
\end{figure*}
\begin{figure*}[htbp]
	\centering
	\includegraphics[width=0.95\textwidth]{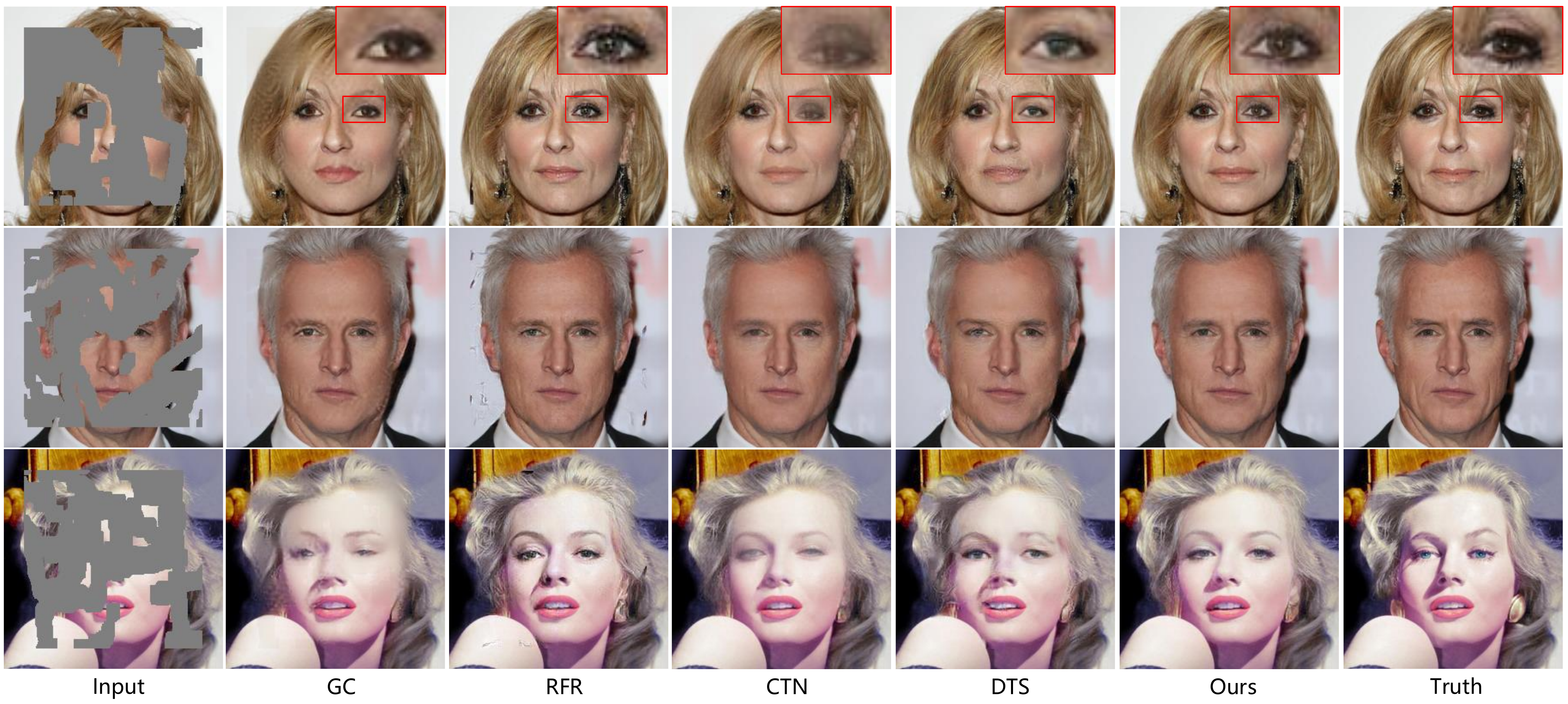} 
	\caption{Qualitative results on the CelebA-HQ with GC \protect\cite{yuGC_2019ICCV}, RFR \protect\cite{Lirfr_2020CVPR}, CTN \protect\cite{dengyemm2021}, DTS \protect\cite{Guodts_2021ICCV} and our $T$-former. (Best viewed with zoom-in)}
	\label{fig:celeba}
\end{figure*}
\begin{figure*}[htbp]
	\centering
	\includegraphics[width=0.95\textwidth]{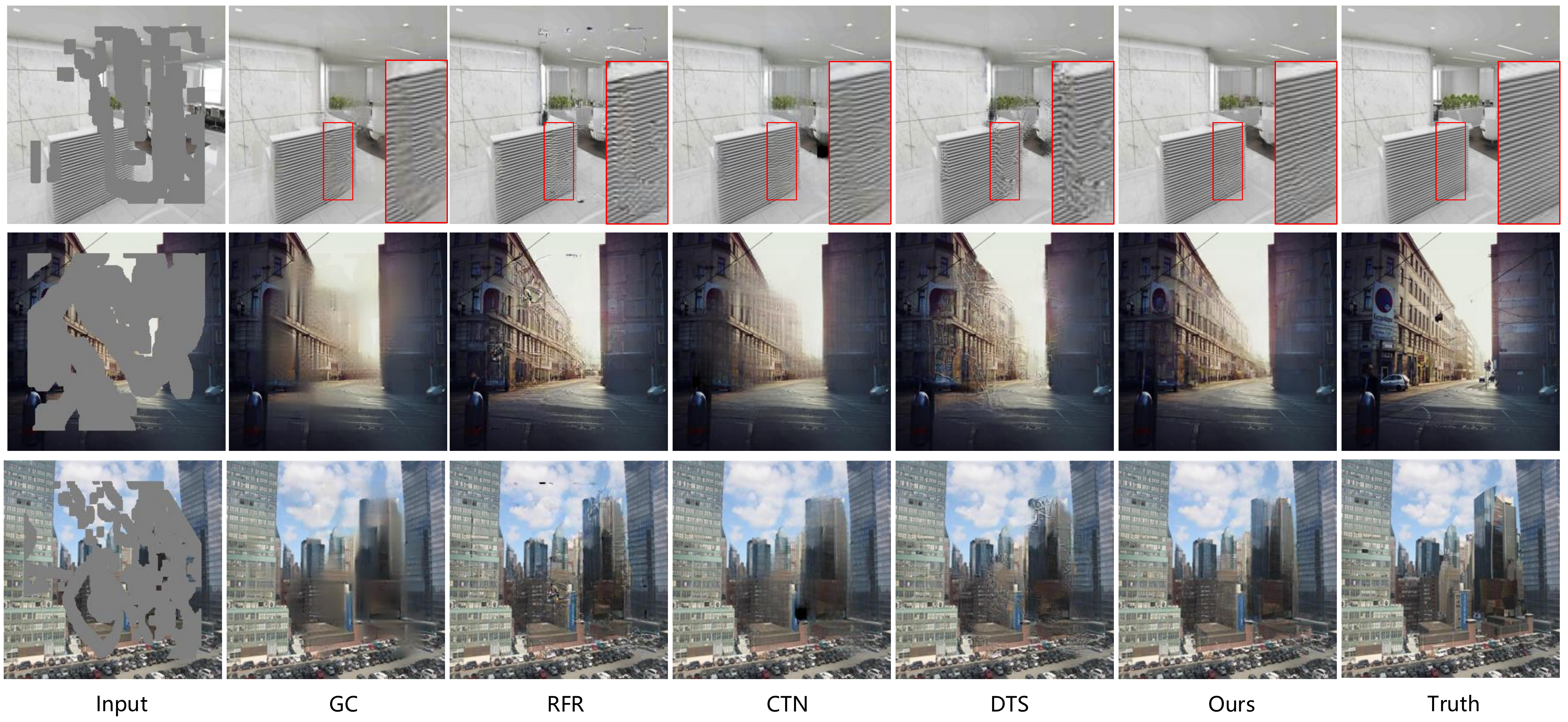} 
	\caption{Qualitative results on the Places2 with GC \protect\cite{yuGC_2019ICCV}, RFR \protect\cite{Lirfr_2020CVPR}, CTN \protect\cite{dengyemm2021}, DTS \protect\cite{Guodts_2021ICCV} and our $T$-former. (Best viewed with zoom-in)}
	\label{fig:pl2}
\end{figure*}

\paragraph{Baselines} 
To demonstrate the effectiveness of our $T$-former, We compare with the following baselines for their state-of-the-art performance:
\begin{itemize}
	\item GC \cite{yuGC_2019ICCV}: a CNNs-based inpainting model that exploits the gating mechanism and the contextual attention~\cite{Yuca_2018CVPR} to get high-quality complementary images..
	\item RFR \cite{Lirfr_2020CVPR}: a recurrent inpainting method with a special contextual attention that recurrently recovers the missing and progressively strengthens the result.
	\item CTN \cite{dengyemm2021}: a transformer-style model for image inpainting relies on a patch-based version of the attention operator to model long-range dependencies between features.
	\item DTS \cite{Guodts_2021ICCV}:  a dual U-net inpainting model based CNNs, which recovers corrupted images by simultaneous modeling structure-constrained texture synthesis and texture-guided structure reconstruction
\end{itemize}

\paragraph{Quantitative Comparison} 
Following previous inpainting works~\cite{dengyemm2021,Guodts_2021ICCV}, we chosen  FID (Fr\'{e}chet Inception Distance)~\cite{NIPS2017_fidgan}, PSNR (peak signal-to-noise ratio), SSIM (structural similarity index) to assess our model. And according to the masks with different masking percentage provided by the dataset~\cite{pc_2018ECCV}, the performance of different models under different damage degrees (mask ratio) is tested in Table~\ref{tab:exp_all}.
In addition, we show the number of parameters and the computational complexity (multiply-accumulate operations, MAC) for each model in Table~\ref{tab:com}.
SSIM and PSNR are widely used in image quality assessment for restoration tasks, quantifying the pixel and structural similarities between pairs of images. In addition we adopt the FID, a generally used numeric metric in the task of image generation, to evaluate the image distribution between the inpainting results and the original images. As shown in Table~\ref{tab:exp_all} and Table~\ref{tab:com}, benefiting from the long-distance dependency capture capability and the dynamic nature of the parameters brought by the proposed linear attention, our $T$-former, in the face of different scenarios (datasets) and encountering different breakage situations, can give relatively good complemented images with a low number of parameters and computational complexity.

\paragraph{Qualitative Comparisons} 
Figures~\ref{fig:paris}, \ref{fig:celeba}, and \ref{fig:pl2} show some comparison results between our and the baseline models on the three data sets Paris \cite{Pathak_2016_CVPR}, CelebA-HQ \cite{pgancelebahq_iclr2018} and Places2~\cite{places2tpami2018} respectively.
From these results, we can find that GC~\cite{yuGC_2019ICCV} is able to complete the basic semantic filling, but the filled position in the image is prone to blurring, especially when filling images with complex patterns, such as the 2nd row in Figure~\ref{fig:paris} and 2nd row in Figure~\ref{fig:pl2}.
The detailed textures of the images complemented by RFR~\cite{Lirfr_2020CVPR} look quite good, but the results are prone to obvious artifacts and are prone to semantic inconsistencies. As in the 1st and 2nd rows of Figure \ref{fig:celeba}, both images generated by RFR show the problem of inconsistent eye color.
CTN~\cite{dengyemm2021} also performs quite well, but its results are occasionally blurred (Figure~\ref{fig:pl2}, line 2) and also prone to black artifacts as shown in (Figure~\ref{fig:pl2}, line 1). DTS~\cite{Guodts_2021ICCV} performs quite well with simple content images, but when it comes to images with complex patterns, the fill content appears to be significantly disorganized, as shown in the 1st row of Figure~\ref{fig:pl2}.
Compared to these baselines, in most cases our complemented images look more reasonable and realistic. 

\begin{table}[htbp]
	\centering
	\caption{Ablation study on the Paris. The $\downarrow$ indicates lower is better, while $\uparrow$ indicates higher is better}
		\resizebox{0.46\textwidth}{!}{
	\begin{tabular}{c|c|cccc}
		\toprule
		\multicolumn{2}{c|}{Mask Ratio} & 10-20\% & 20-30\% & 30-40\% & 40-50\% \\
		\midrule
		\multirow{4}[2]{*}{FID$\downarrow$} & w/o $\mathcal{V}$ & 12.20 & 23.93 & 35.14 & 47.99 \\
		& w/o $\mathcal{G}$ & 12.74 & 22.89 & 36.08 & 47.48 \\
		& w/o $\mathcal{G}$+$\mathcal{V}$ & 12.94 & 24.19 & 37.08 & 48.47 \\
		& Ours  & \textbf{12.15} & \textbf{22.63} & \textbf{34.47} & \textbf{46.60} \\
		\midrule
		\multirow{4}[2]{*}{PSNR$\uparrow$} & w/o $\mathcal{V}$ & 32.74 & 29.69 & 27.35 & 25.36 \\
		& w/o $\mathcal{G}$ & 32.72 & 29.68 & 27.36 & 25.33 \\
		& w/o $\mathcal{G}$+$\mathcal{V}$ & 32.70 & 29.66 & 27.33 & 25.28 \\
		& Ours  & \textbf{32.79} & \textbf{29.72} & \textbf{27.47} & \textbf{25.47} \\
		\midrule
		\multirow{4}[2]{*}{SSIM$\uparrow$} & w/o $\mathcal{V}$ & 0.962 & 0.932 & 0.884 & 0.823 \\
		& w/o $\mathcal{G}$ & 0.963 & 0.931 & 0.884 & 0.823 \\
		& w/o $\mathcal{G}$+$\mathcal{V}$ & 0.961 & 0.930 & 0.883 & 0.821 \\
		& Ours  & \textbf{0.964} & \textbf{0.933} & \textbf{0.887} & \textbf{0.825} \\
		\bottomrule
	\end{tabular}}%
	\label{tab:exp:abl}%
\end{table}%

\subsection{Ablation Study}
We analyze the effectiveness of our proposed module. And all the ablation experiments are conducted on the Paris street view. In the ablation experiments we explored two main components: (1) for the effect of the residual-like connection resulting from $V+$ in Eq.~(\ref{eq:matrx}) (or Eq.~(\ref{eq:linears})), i.e. $1$ in $ e \approx 1+x$ in the Taylor expansion, denoted by $\mathcal{V}$. When $\mathcal{V}$ does not exist, our linear attention is somewhat similar to the current implementation of a series of linear attentions~\cite{peng2021random,zhen2022cosformer,pmlr-v119-katharopoulos20a} in the field of natural language processing where softmax operators are replaced by kernel functions.
And the $\mathcal{V}$ is equivalent to adding a new residual term to this family of linear attention operators;
 (2) for the effect of the gating mechanism on the model performance, denoted by $\mathcal{G}$. It can be noticed that both components have a positive impact on the inpainting task. A more interesting point is that it can be found that $\mathcal{V}$ acts more significantly when the input image is more broken, while the effect of $\mathcal{G}$ is independent of the degree of input image breakage. The paper~\cite{NIPS2016_resembe} has showed that the residual connections can be seen as an ensemble of the model, and one more connection is equivalent to one more sub-network. We speculate that when the model encounters difficult scenes (more broken parts of the input image), more sub-networks (with $\mathcal{V}$) are needed to assistant the model get the proper content to fill the missing areas.

\section{Conclusion}
In this paper, we propose $T$-former, a U-net style network built by the proposed linear attention for for image inpainting.
To address the problem that CNNs-based inpainting networks have insufficient long-range modeling capability and the standard self-attention operator has high computational load, we propose a linear attention operator based on Taylor's formula that captures the long-range dependence between features at a small computational cost. In addition, we utilize a gating mechanism to enhance the performance of the proposed linear attentional operator.
Quantitative and qualitative results demonstrate that our proposed $T$-former outperforms state-of-the-art methods in terms of performance and also maintains a relatively small complexity.

\begin{acks}
	This work is jointly supported by the National Key Research and Development Program of China under Grant No. 2017YFA0700800, the General Program of China Postdoctoral Science Foundation under Grant No. 2020M683490, and the Youth program of Shaanxi Natural Science Foundation under Grant No. 2021JQ-054.
\end{acks}

\bibliographystyle{ACM-Reference-Format}
\balance
\bibliography{sample-base}

\end{document}